\documentclass[11pt,letterpaper]{article}
\usepackage{cogsys}
\usepackage[T1]{fontenc}
\usepackage{times}
\usepackage[pdftex]{graphicx} 

\usepackage{wrapfig}
\usepackage[font=small]{caption}
\usepackage{dialogue}
\usepackage{amsmath}
\usepackage{amssymb}
\DeclareMathOperator*{\argmin}{argmin}

\usepackage[shortlabels]{enumitem}
\usepackage{algorithm2e}
\usepackage{ragged2e}
\DeclareCaptionFormat{linebreak}{#1#2#3}
\usepackage{xcolor}

\usepackage{hyperref}
\usepackage[export]{adjustbox}

\usepackage{natbib}
\cogsysheading{X}{20XX}{1-6}{X/20XX}{X/20XX}

\ShortHeadings{Neurosymbolic AI for Situated Language Understanding}
              {N.\ Krishnaswamy, J.\ Pustejovsky}

\begin{document}

{\obeyspaces\gdef {\ }}
\global\newbox\codebox
\global\newbox\savedcodebox
\gdef\sverbatim{\bgroup\def\endsverbatim{\egroup\egroup\egroup\mbox{\box\codebo\
x}}\def\savecode{\egroup\egroup\egroup\global\setbox\savedcodebox\copy\codebox}\
\def\par{\egroup\vspace{-0.3em}\hbox\bgroup}\tt\obeylines\obeyspaces\global\set\
box\codebox\vbox\bgroup\hbox\bgroup}
\gdef\savedcode{\copy\savedcodebox}

\newcommand{\avm}[1]{{\setlength{\arraycolsep}{0.8mm}
                       \renewcommand{\arraystretch}{1.2}
                       \left[
                       \begin{array}{l}
                       \\[-4mm] #1 \\[-4mm] \\
                       \end{array}
                       \right]
                    }}
\newcommand{\avmplus}[1]{{\setlength{\arraycolsep}{0.8mm}
                       \renewcommand{\arraystretch}{1.2}
                       \left[
                       \begin{array}{l}
                       \\[-2mm] #1 \\[-2mm] \\
                       \end{array}
                       \right]
                    }}
\newcommand{\att}[1]{{\mbox{\small {\bf #1}}}}
\newcommand{\attval}[2]{{\mbox{\small {\sc #1}}\ =\ {{#2}}}}
\newcommand{\attvalterm}[2]{{\mbox{\small {\sc #1}}\ =\ {\myvalue{#2}}}}
\newcommand{\attvaltyp}[2]{{\mbox{\small{\sc #1}}\ =\ {\myvaluebold{#2}}}}
\newcommand{\myvalue}[1]{{\mbox{\normalsize {\it #1}}}}
\newcommand{\myvaluebold}[1]{{\mbox{\small {\bf #1}}}}
\newcommand{\ind}[1]{{\setlength{\fboxsep}{0.5mm} \: \fbox{{\tiny #1}} \:}}

\newcommand{\attnoval}[1]{{\mbox{\small {\sc #1}}  }}
\newcommand{\attvaltypnoatt}[2]{{\mbox{$\;\;\;\;\;\;\;\;\;\;$} \  \ {\myvaluebold{#2}}}}
\def\implies{\Rightarrow}

\newcounter{lingex}

\newenvironment{enumerate*}%
  {\begin{enumerate}%
    \setlength{\itemsep}{0pt}%
    \setlength{\parskip}{0pt}}%
  {\end{enumerate}}

\newenvironment{itemize*}%
  {\begin{itemize}%
    \setlength{\itemsep}{0pt}%
    \setlength{\parskip}{0pt}}%
  {\end{itemize}}

\title{Neurosymbolic AI for Situated Language Understanding}
 
\author{Nikhil Krishnaswamy}{nkrishna@colostate.edu}
\address{Department of Computer Science, Colorado State University,
	Fort Collins, CO USA \\
	Department of Computer Science, Brandeis University,
	Waltham, MA USA}
\author{James Pustejovsky}{jamesp@brandeis.edu}
\address{Department of Computer Science, Brandeis University,
	Waltham, MA USA}
\vskip 0.2in

\begin{abstract}
In recent years, data-intensive AI, particularly the domain of natural language processing and understanding, has seen significant progress driven by the advent of large datasets and deep neural networks that have sidelined more classic AI approaches to the field.  These systems can apparently demonstrate sophisticated linguistic understanding or generation capabilities, but often fail to transfer their skills to situations they have not encountered before.  We argue that computational {\it situated grounding} provides a solution to some of these learning challenges by creating situational representations that both serve as a formal model of the salient phenomena, and contain rich amounts of exploitable, task-appropriate data for training new, flexible computational models.  Our model reincorporates some ideas of classic AI into a framework of {\it neurosymbolic intelligence}, using multimodal contextual modeling of interactive situations, events, and object properties.  We discuss how situated grounding provides diverse data and multiple levels of modeling for a variety of AI learning challenges, including learning how to interact with object affordances, learning semantics for novel structures and configurations, and transferring such learned knowledge to new objects and situations.
\end{abstract}

\section{Introduction}
\label{sec:intro}

Over the past fifteen to twenty years, AI has seen remarkable growth, from a bust beset by disillusionment with unmet expectations to a behemoth at the center of modern computer science, and a maturing set of technologies to match \citep{menzies2003guest, mccarthy2007here, liu2018artificial}.  A significant proportion of this growth has been driven by advances in natural language processing (NLP), previously a difficult problem with brittle solutions, and now a mainstay of technologies in everyday use.  Developers without substantial prior knowledge of AI or linguistics can now use robust pipelines for natural language tasks such as tokenization, parsing, or speech recognition.  Within the last decade, the 2010s, NLP progress was kicked into overdrive, largely due to the developments in deep learning and the concurrent emergence of large datasets and affordable GPUs for processing them.  Deep learning has been applied to tasks such as question answering \citep{sultana2020review}, dialogue systems \citep{zaib2020short}, and text generation \citep{iqbal2020survey}.

Many of the biggest successes in NLP have been driven by large, pre-trained, task-agnostic language models, such as ELMo \citep{peters2018deep}, BERT \citep{devlin2018bert}, and the GPT family \citep{radford2019language}.  These language models lend themselves well to transfer learning with task-specific fine tuning, and facilitate the generation of text that is grammatical, largely coherent, and usually on-topic given an initial prompt.  They are also simple to deploy and well-pipelined for general use in larger applications or just as a demonstration of capability.  An example is given below, generated using the GPT-2 language model, a transformer based language model \citep{vaswani2017attention} trained over 40GB of text from the internet.  The input is the first sentence of the \href{https://en.wikipedia.org/wiki/Natural_language_processing}{Wikipedia article on NLP}, given in bold, and continued by the language model.

\leftskip1cm\relax
\rightskip1cm\relax
\textit{\textbf{Natural language processing (NLP) is a subfield of linguistics, computer science, information engineering, and artificial intelligence concerned with the interactions between computers and human (natural) languages, in particular how to program computers to process and analyze large amounts of natural language data.} NLP encompasses methods to manipulate and categorize textual data and information in ways similar to biological organisms.  As such, NLP is an excellent use case for learning intelligent bots, as you can teach a computer how to process natural language data in order to explore, build, test, validate, and apply advanced learning methods in response to the data [...]}

\leftskip0cm\relax
\rightskip0cm\relax
The generated text is largely syntactically and semantically correct and mostly on topic, with a few instances of questionable phrasing (e.g., ``learning intelligent bots'') or possibly off-topic content or factual inaccuracies (e.g., references to categorizing language like biological information).  The system is talking about NLP as one might expect a non-expert to, the errors it makes are similar to those a human layperson might make, it incorporates a significant amount of contextual information in the generated sample, and it is clear that NLP systems have come a long way on this task from the days of finite state automata and simple N-gram models.

Nonetheless, despite the apparent success of language models in NLP tasks, there are a number of ways in which they fail to demonstrate intelligence or understanding as commonly defined, in particular on tasks that would typically be trivial for a human.  In previous work \citep{krishnaswamy2019situated,mcneely2019user}, we have given the example of asking a computer or smartphone the simple (for a human) question ``What am I pointing at?''  Put simply, an NLP system such as one embedded in a smartphone or smart-speaker lacks the machinery to answer the question, being unable to see you or the surrounding context you and the device inhabit, and so tends to dodge the question\footnote{Try it with your device! Answers may include ``Interesting question'' (Siri) or ``You're a great person to chat with!'' (Google Mini).}.  Large predictive language models fare little better, as shown below.

\leftskip1cm\relax
\rightskip1cm\relax
\textit{\textbf{Q: What am I pointing at?}}

\textit{A: There was nothing I could detect at the start of the game but not many areas. It took some training to find my way around but by the end of the game I had discovered the best way to move and how to take advantage of my surroundings [...]}

\leftskip0cm\relax
\rightskip0cm\relax
Although trained on a enormous amount of text, these models lack knowledge of the current situational context, because that context is supplied using non-textual modalities, and so the main advertised advantage of a pre-trained language model---the ability to transfer knowledge learned from observed text to previously unencountered text---disappears.  We now have many usable interactive systems, such as smartphones and the whole internet-of-things, but the large datasets and compute power that facilitate high-performing NLP fail in many contexts in which we might wish to use these devices, and might expect them to function as if they truly understand us.  Put simply, the current state of the technology runs up against a wall because these systems exist in a situated context (a home, an office, a car, a pocket, etc.), but lack the ability to validate information across the different modalities of description that might be implicated in all these situations.  They also lack background knowledge about other entities present in the situation.  Therefore, how can we expect to interface with these devices when something so basic to a human---like ``What am I pointing at?''---fails?

In this paper, we will discuss our {\it situated grounding} approach to multimodally encoding context, and our platform, {\it VoxWorld}, which demonstrates real-time modeling of context through multimodal grounding of object and event properties in a simulation environment, and the {\it common ground} that arises between interlocutors in the course of an interaction.  We will demonstrate how situated grounding methods within VoxWorld provide diverse types of data suited to a number of different learning challenges within AI, and how deploying this data and their associated models within a {\it neurosymbolic} intelligence framework addresses three novel challenges in AI: affordance learning, structure/configuration learning, and transfer learning.

\section{Multimodal Communication in Context}
\label{sec:mcic}

As sophisticated as current task-based AI systems are and as intelligent as they can behave in their domains, they often fail in understanding and communicating crucial information about their situations.  Robust communicative interaction between humans and computers requires that:
\begin{enumerate}
\item All parties must be able to recognize input and generate output within multiple modalities appropriate to the context (e.g., language, gesture, images, actions, etc.);
\item All parties must demonstrate understanding of contextual grounding and the space in which the conversation takes place (e.g., co-situated in the same space, mediated through an interface, entirely disconnected, etc.);
\item All parties must appreciate the consequences of actions taken throughout the dialogue.
\end{enumerate}

Central to all these is the notion of {\it semantically grounding} a concept to a situation.  Importantly, certain modalities are better suited to grounding certain kinds of information than others (for example, deictic gesture---i.e., pointing---grounds naturally to locations, while language may be better at grounding concept labels or attribute descriptions).  Nonetheless, ``grounding'' in currently-practiced NLP typically refers to kinds of multimodal {\it linking}, such as semantic roles to entities in an image \citep{yatskarsituation}, or joint linguistic-visual attention between a caption and an image \citep{li2019visualbert}.

This type of annotation and training on large quantities of multimodal data resembles a cross-modal linking equivalent of the traits that have made large language models successful at their tasks: they are annotated, they contain structured data, and they contain mechanisms for extracting many sophisticated linguistic features, even if they have no built-in understanding of linguistic structure.  In fact, they resemble the classic NLP pipeline but on a much larger scale \citep{tenney2019bert}.  When applied to multimodal data, however, and pertaining to the representation of context, the same classic NLP pipeline (or deep-learned equivalent) fails to work as well.

Multimodal tasks rely on the contexts established between and across modalities \citep{matuszek2018grounded}, and so we propose that the difficulties faced by multimodal end-to-end systems, as well as the difficulty evaluating the state of the task is largely because contextual encoding still tends to be hit-or-miss, and the nature of the analytic and structural units of context, as humans use for sensitive contextual reasoning, remain the subjects of debate.

Nevertheless, human reasoning is sensitive to contextual modeling, and methods of contextual modeling in AI have followed the field from logical-symbolic models of context (``good old-fashioned AI'' or {\it GOFAI}) before the AI winter of the 1980s to probabilistic and deep-learned vector similarity in the 2010s.  Recently, GOFAI methods have resurfaced in the increasingly machine learning-driven modern AI community as a method of reincorporating some of the structure they provide into the flexible representations provided by deep learning (e.g., \cite{besold2017neural,garcez2019neural,mao2019neuro,marcus2019rebooting}.\footnote{As well as keynote addresses given at AAAI 2020 by David Cox of IBM, Henry Kautz of the University of Rochester, and Turing Award winners Geoffrey Hinton, Yann LeCun, and Yoshua Bengio.}  The question of better incorporating contextual structure into deep learning necessarily raises the question of the analytic and structural units of context.

Following on \cite{clark1983common}, \cite{asher1998common}, \cite{stalnaker2002common}, \cite{tomasello2007shared}, \cite{abbott2008presuppositions}, \cite{pustejovsky2018actions}, and others, we have previously proposed the notion of a {\it computational common ground} that emerges between interlocutors as they interact, and facilitates further communication by providing common knowledge among agents \citep{pustejovsky2017creating}.  Common ground is one such method of encoding and analyzing situational and conversational context.

\begin{figure}[h!]
\centering
\begin{tabular}{cc}
    \centering
  	\includegraphics[width=1.6in,valign=m]{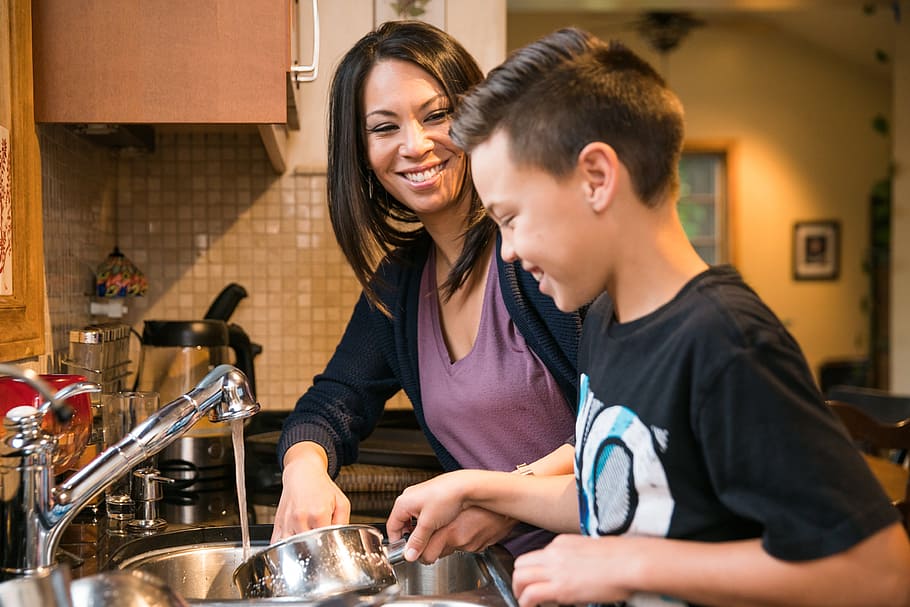}
	&
    \centering
\begin{tabular}{|l|l|}
    \hline
        {\bf Entity Type} & {\bf Examples} \\
    \hline
        Agents & ``you,'' ``I,'' ``us,'' etc. \\
    \hline
        Beliefs, desires, & {\bf Know}$_{mother}$smile(son), etc.
        \\
        intentions & ``Goals under Discussion''\\
    \hline
        Objects & cups, plates, knives, ``it,'' ``them,'' etc. \\
    \hline
        Space & $\mathcal{E}$ (Embedding space) \\
    \hline
    \end{tabular}
    \end{tabular}
\caption{Two humans interacting in a shared task with example common ground entities.}
\label{fig:interaction-cg}
\end{figure}

We break down computational common ground into representations of: the agents interacting; their beliefs, desires and intentions (BDI); the objects involved in the interaction; and the minimal embedding space $\mathcal{E}$ required to execute the activities implicated during the course of the interaction.  All these parameters also include the terms used to discuss them.  For instance, in Fig.~\ref{fig:interaction-cg}, the mother and son agree that they share a goal to, e.g., put the dishes away, empty the sink, clean the dishes, etc. (if one of them does not share this belief, this impacts the way both of them will communicate about the task and their beliefs about it).  This in turn implicates the properties of the objects involved, e.g., what it means to have a clean plate vs. a dirty plate with relation to what a plate is for.

Object properties are a topic of much discussion in semantics, including Generative Lexicon theory \citep{Pustejovsky1995,pustejovsky2019lexicon}, and are also of interest to the robotics community \citep{dzifcak2009and}.  Object properties, though important for theoretical semantics and practical applications of modern intelligent systems, pose a problem for even some of the most sophisticated task-based AI systems.  A formal structure provided by the elements of common ground and situational context proposes a possible solution to these difficulties.  Subsequently, we detail experiments we have been conducting in VoxWorld, the situated grounding platform based on the VoxML modeling language \citep{PUSTEJOVSKY16.1101}.  These experiments combine neural learning and symbolic reasoning approaches to address affordance learning, structure learning, and transfer learning for an intelligent agent.

\section{Modeling Context}
\label{sec:modeling}

Object properties and the events they facilitate \citep{gibson1977} are a primary component of situational context.  Gibson's initial formulation of affordances vaguely defines the term as what the environment ``offers the animal.''  Gibson refers to the term as ``something that refers to both the environment and the animal in a way that no existing term does.  It implies the complementarity of the animal and the environment'' \citep{gibson1979}.

We use the term in our work in a way that attempts to cover the extensive ground that Gibson uses it for, while maintaining a clear relation between the environment (including object configuration as a positioning, or {\it habitat}), the properties of an object that allows it to be used for certain behaviors (e.g., the ``graspability'' of a handle), and the language used to describe these behaviors and ground them to an environment or situation, as has been explored in recent neural AI work (e.g., \cite{hermann2017grounded,das2017embodied}),

For instance, in Fig.~\ref{fig:cups}, the cup on the left is in a position to be {\it slid} across its supporting surface while the cup on the right is in a position to be {\it rolled}.  Executing one or the other of these actions would require the cup to be placed in a prerequisite orientation, and may result in concomitant effects, such as anything contained in the cup spilling out (or not).

\begin{wrapfigure}{l}{.35\textwidth}
\vspace{-6mm}
\begin{center}
\includegraphics[width=.15\textwidth]{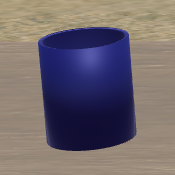}
\includegraphics[width=.15\textwidth]{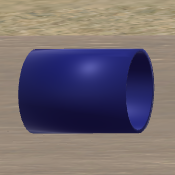}
\caption{\label{fig:cups}Cups in different habitats.}
\end{center}
\vspace{-8mm}
\end{wrapfigure}

We correlate these {\it afforded} behaviors (a la Gibson, and \cite{Pustejovsky1995}'s {\it telic} roles) with the notion of habitats \citep{pustejovsky2013dynamic,mcdonald2013representation}, or conditioning environments that facilitate affordances.  Formally, we capture these properties in the modeling language VoxML \citep{PUSTEJOVSKY16.1101}, that captures common object and event semantics, with a particular focus on habitats and affordances.  VoxML models ontological information that is difficult to learn from corpora due to being so common that it is rarely documented and therefore not available to machine learning algorithms.  VoxML provides the format for the symbolic encodings of our neurosymbolic pipeline.  Each component of a VoxML encoding (e.g., object shape, event semantic class, individual habitat, affordance, etc.) can be hand-encoded, extracted from corpora, or learned, providing a way to habituate common qualitative knowledge into a structured but flexible representation.  This qualitative knowledge is important to reflect human-like qualitative reasoning capabilities in a computational context.  When reasoning about a ball rolling, humans do not need to know the exact value of parameters like speed or direction of motion, but to simulate the event computationally, every variable must have a value for the program to run.  VoxML provides a structured encoding of properties for these variables that allows a system to generate values when needed.  Fig.~\ref{fig:cup-voxml} shows the VoxML encoding for a cup.  Note the intrinsic upward orientation of the habitat $H_{[3]}$ where the cup's Y-axis is aligned with that of the world, and the afforded behaviors that may be conditioned on a particular habitat, or may be available in any habitat (denoted $H\rightarrow$).

\begin{figure}[h!]
\centering
\tiny\def\baselinestretch{1.1}
$\avmplus{\att{cup}\\
	\attval{lex}{\avmplus{
	    \attvaltyp{pred}{cup}\\
	    \attvaltyp{type}{physobj$\bullet$artifact}
	}}\\
	\attval{type}{\avmplus{
		\attvaltyp{head}{cylindroid[1]}\\
		\attvaltyp{components}{surface[1], interior[2]}\\
		\attvaltyp{concavity}{concave[2]}\\
		\attvaltyp{rotat\_sym}{Y}\\
		\attvaltyp{refl\_sym}{XY, YZ}
	}}\\
	\attval{habitat}{\avmplus{
        \attval{Intr}{[3]\avmplus{
            \attvaltyp{up}{$align(\bar{Y},\mathcal{E}_Y)$}\\
            \attvaltyp{top}{$top(+Y)$}
        }}\\
		\attval{Extr}{\avmplus{
		    {[4]\avmplus{
                \attvaltyp{up}{$align(\bar{Y},\mathcal{E}_Y)$}\\
                \attvaltyp{top}{$top(-Y)$}
            }}\\
            {[5]\avmplus{
                \attvaltyp{up}{$align(\bar{Y},\mathcal{E}_{\perp\-Y})$}
            }}
        }} 
	}}\\
	\attval{afford\_str}{\avmplus{
		\attvaltyp{A$_1$}{$H \rightarrow [put(x, y, on([1])]support([1], y)$}\\
		\attvaltyp{A$_2$}{$H_{[3]} \rightarrow [put(x, y, in([2])]contain([2], y)$}\\
	    \attvaltyp{A$_3$}{$H \rightarrow [grasp(x, [1])]hold(x, [1])$}\\
        \attvaltyp{A$_4$}{$H \rightarrow [lift(x, [1])]hold(x, [1])$}\\
	    \attvaltyp{A$_5$}{$H \rightarrow [ungrasp(x, [1])]release(x, [1])$}\\
	    \attvaltyp{A$_6$}{$H_{[3,4]} \rightarrow [slide(x, [1])]\mathcal{R}$}\\
		\attvaltyp{A$_7$}{$H_{[5]} \rightarrow [roll(x, [1])]\mathcal{R}$}\\
		\attvaltypnoatt{\;}{...}
	}}\\
	\attval{embodiment}{\avmplus{
	    \attvaltyp{scale}{$<$agent}\\
	    \attvaltyp{movable}{true}
	}}
}$
\def\baselinestretch{1.9}
    \caption{VoxML encoding for a [[{\sc cup}]] voxeme.}
    \label{fig:cup-voxml}
\end{figure}

\subsection{Multimodal Simulations}
\label{ssec:simulations}

The situated, simulated environments of the VoxWorld platform bring together three notions of simulation from computer science and cognitive science \citep{pustejovsky2019situational}:

\begin{enumerate}
\item {\it Computational simulation modeling}. That is, variables in a model are set and the model is run, such that the consequences of all possible computable configurations become known. Examples of such simulations include models of  climate change, the tensile strength of materials,  models of biological pathways, and so on. The goal is to arrive at the best model by using simulation techniques. 

\item {\it Situated embodied simulations}, where the agent is embodied with a dynamic point-of-view or avatar in a virtual or simulated world.  Such simulations are used for training humans in scenarios such as flight simulators or combat situations, and of course are used in video gaming as well.  In these contexts, the virtual worlds assume an embodiment of the agent in the environment, either as a first-person restricted POV or an omniscient movable embodied perspective.  The goal is to simulate an agent operating within a situation. 

\item {\it Embodied theories of mind}.  \cite{craik1943nature} and, later, \cite{johnson1987could} develop the notion that agents carry a mental model of external reality in their heads.   \cite{johnson2002conditionals} represent this model as a situational possibility, capturing what is common to different ways the situation may
occur.  Simulation Theory in philosophy of mind focuses on the role of ``mind reading'' in modeling the representations and communications of other agents \citep{gordon1986folk,goldman1989interpretation,heal1996simulation,goldman2006simulating}. Simulation semantics (as adopted within cognitive linguistics and  practiced by \cite{feldman2010embodied}, \cite{narayanan2010mind}, \cite{bergen2012louder}, and \cite{evans2013language}) argues that language comprehension is accomplished by such mind reading operations. There is also an established body of work within psychology arguing for {\it mental simulations} of future or possible outcomes, as well of perceptual input \citep{graesser1994constructing,barsalou1999perceptions,zwaan1998situation,zwaan2012revisiting}.  The goal is semantic interpretation of an expression by means of a simulation, which is either mental (a la Bergen and  Evans) or interpreted graphs such as Petri Nets (a la Narayanan and Feldman).
\end{enumerate}

\cite{krishnaswamy2017montecarlo} brings the model testing of (1), the situated embodiment of (2), and the modeling machinery of (3) together into Monte-Carlo visual simulation of underspecified motion predicates, which forms the backbone of a situated approach to learning and language understanding.  Given a label (symbol) of a motion verb, there may be a large space of potential specific instantiations of that motion that satisfy the label.  The specifics may depend on the objects involved, and may contain many underspecified variable values (e.g., speed of motion, exact path---depending on the verb, etc.).

In an interaction, as in Fig.~\ref{fig:interaction-cg}, each agent maintains their own model of what the other agent knows, including respective interpretations of vocabulary items.  For instance, if the mother says ``pass me that plate'' and the son throws it at her, it becomes clear to her that his interpretation of ``pass'' differs from hers.  Since the computer system operationalizes all these motion predicates in terms of primitive motions like {\it translate} and {\it rotate}, it needs a model that accommodates flexible representations of these primitive motions and of their composition into more complex motions.

\begin{figure}[h!]
\centering
\includegraphics[width=.75\textwidth]{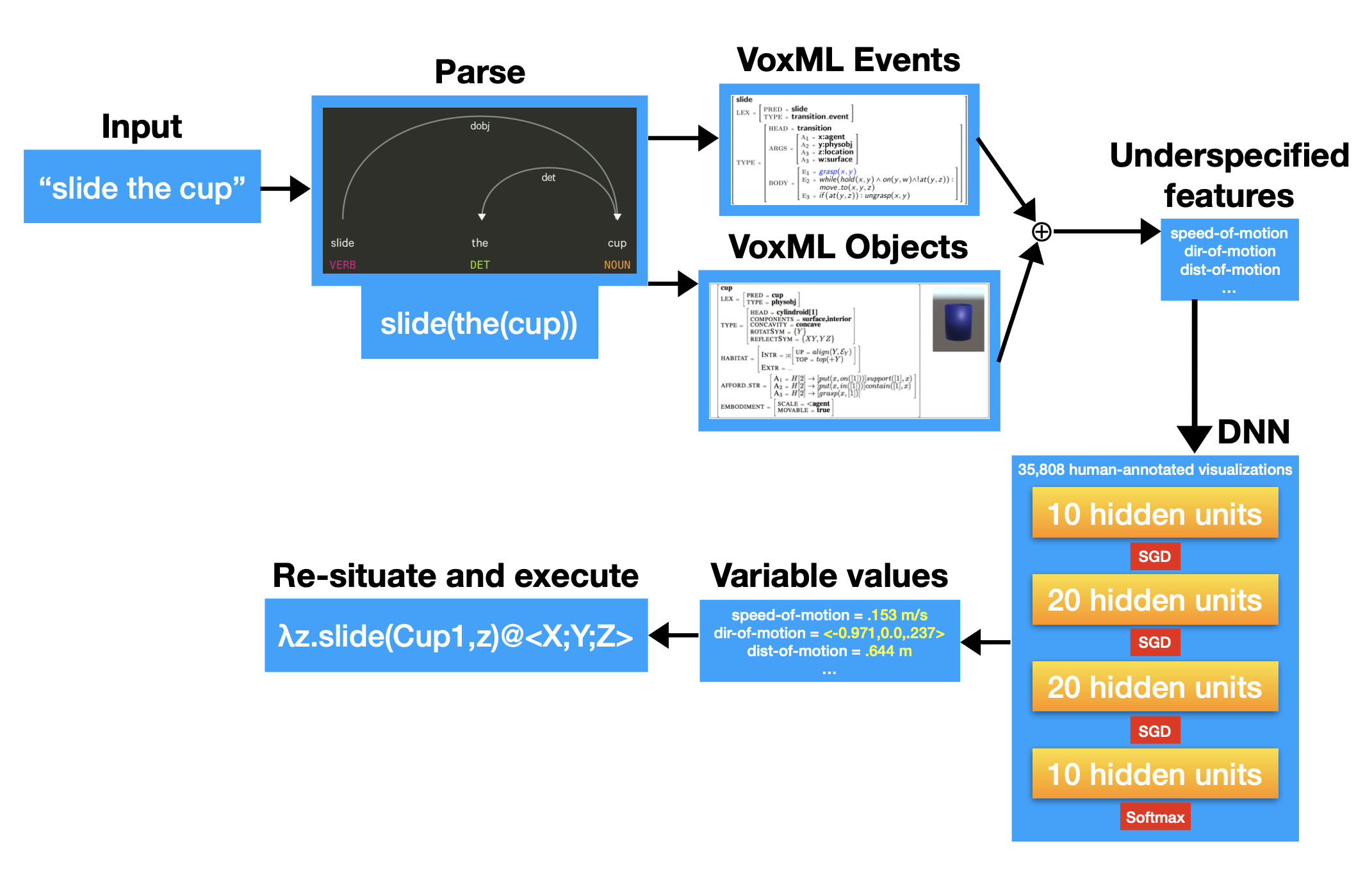}
\caption{Neurosymbolic pipeline for generating multimodal simulations.}
\label{fig:mmsim}
\end{figure}

The Monte-Carlo simulation approach of VoxWorld provides the model on which to operationalize these complex motion predicates in ways that behave according to the preconceived notions of a typical human user.  Given an input (a simple event description in English), the input is parsed and broken out into VoxML representations of the objects, events, and relations involved.  These individual structured representations are then {\it recomposed}.  From that recomposition, the variables of the composed representation that remain unassigned are extracted as the underspecified features.

The VoxML- and Unity-based VoxSim software \citep{krishnaswamy2016voxsim} was then used to generate over 35,000 animated visualizations of a variety of common motion events (put, slide, lift, roll, lean, etc.) with a vocabulary of common objects (cups, pencils, plates, books, etc.), that displayed a wide variety of underspecified variables in their respective operationalizations.  Each visualization was given to 8 annotators each, along with two other variant visualizations of the same input event, and the annotators were asked to choose the best one, as well as to choose the best event caption for each visualization\footnote{Data is available at \href{https://github.com/nkrishnaswamy/underspecification-tests}{https://github.com/nkrishnaswamy/underspecification-tests}}.  We then extracted the range of values assigned to underspecified parameters in those visualizations which annotators judged appropriate, and used a feedforward deep neural network (DNN) to predict the best values for underspecified parameters given an event input in plain English.  When given an input text, VoxSim runs the underspecified parameter symbols through the model, and the resultant output values are assigned to the relevant input parameters, resituated in the scene, and executed in real time to create an appropriate visualization of the input event.  Fig.~\ref{fig:lean-cup} shows the resulting state for one such visualization for ``lean the cup on the book''.

\begin{wrapfigure}{r}{.4\textwidth}
\vspace{-6mm}
\begin{center}
\includegraphics[width=.4\textwidth]{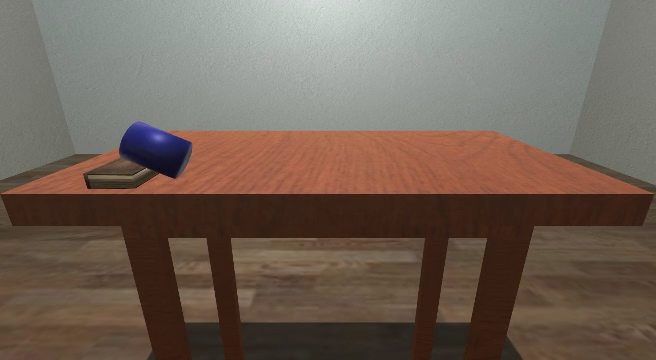}
\caption{\label{fig:lean-cup}Visualization of ``lean the cup on the book.''}
\end{center}
\vspace{-16mm}
\end{wrapfigure}

This pipeline is shown in Fig.~\ref{fig:mmsim} and serves as the basis for interactively exploring learning and reasoning through situated grounding.

\section{Interactive Learning of Object \\ Affordances}
\label{sec:affordances}

Underspecified parameters in a predicate can also be inferred from the properties of objects, namely the habitats which they can occupy and the behaviors afforded by them.  For instance, if a cup is both {\it concave} and symmetric around the {\it Y-axis}, then there is no need to explicitly specify the orientation of the concavity; we can infer that it is aligned with the object's Y-axis, and this in turn requires that certain conditions (habitats) be enforced for certain behaviors (affordances) to be taken advantage of, such as putting something in the cup, or grasping the cup appropriately in order to drink from it \citep{krishnaswamy2016multimodal}.

Previously, we relied on hand-crafted object affordance encodings in VoxML (e.g., see Fig.~\ref{fig:cup-voxml}), but this is an inefficient process and hard to scale.  To automatically explore affordances such as {\it grasping}, a system must have an agent capable of grasping items, namely an {\it embodied, situated agent} that explores situated grounding from its own dynamic point of view.  In \cite{krishnaswamy2017communicating} and \cite{narayana2018cooperating}, we examined the problem of situatedness and communication within a situated context in an encounter between two ``people'': an avatar modeling multimodal dialogue with a human.

\begin{figure}[h!]
\begin{center}
\includegraphics[width=.5\textwidth]{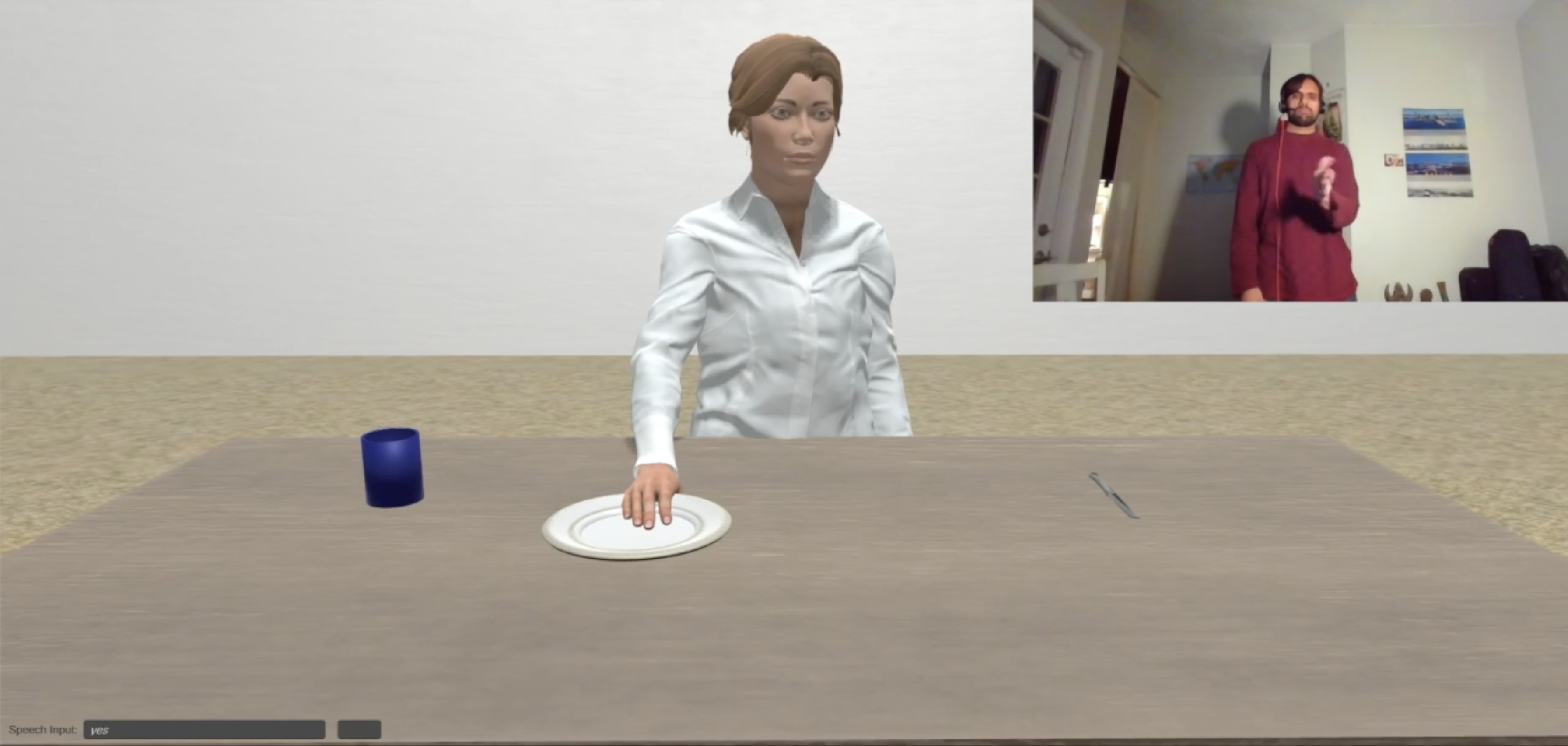}
\caption{\label{fig:diana}Diana interacting with a human.}
\end{center}
\end{figure}

Our agent in VoxWorld, known as Diana, is situated in a virtual VoxSim environment, consumes input from 3rd party or custom speech recognition, and can see her human interlocutor's gestures with custom recognition algorithms running on deep convolutional neural networks trained on over 8 hours of annotated video and depth data from a Microsoft Kinect\textsuperscript{TM}.  The human can gesture about objects in Diana's virtual world, making Diana an interactive collaborator.\footnote{A full-length video demo of Diana can be viewed \href{http://www.voxicon.net/wp-content/uploads/2020/04/Diana-AAAI-2020-Demo-BGVideo.mp4}{here}.}

A major challenge for collaborative AI is data availability for different tasks.  Nonetheless, we can use the situated grounding environment of VoxWorld and the interactive mechanisms enabled by an agent like Diana to access detailed, multimodal data for a variety of tasks. Diana's default vocabulary of 34 gestures contains a downward-opening ``claw'' gesture used to mean {\it  grasp}.  This gesture is sufficient to signal grasping an object like a block.  However, in Diana's ``kitchen world'' scenario, containing common household objects, she comes across items, like plates, that cannot be grasped in this way.  In that case, she must estimate positions on the object where it is graspable.

{\it Grasp point inference} uses the symmetry of objects as encoded in VoxML.  Objects have rotational and reflectional symmetry, such that a cup has rotational symmetry around its Y-axis and reflectional symmetry across its XY- and YZ-planes, while a knife has only reflectional symmetry across its YZ-plane in default orientation.  

\begin{wrapfigure}{l}{.5\textwidth}
\vspace{-6mm}
\begin{center}
\includegraphics[height=.9in]{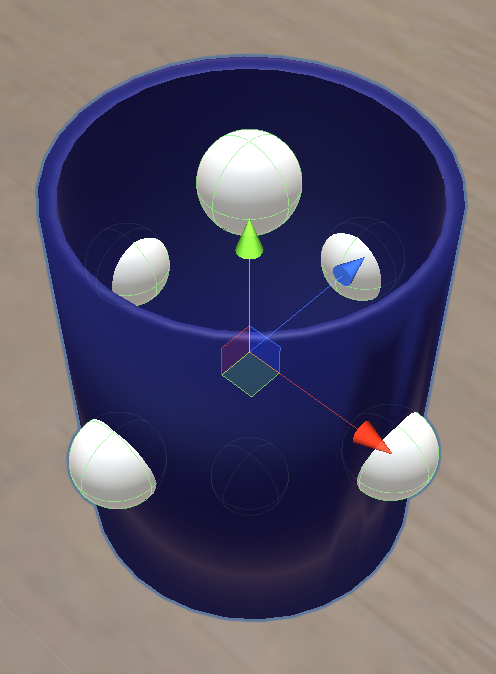}
\includegraphics[height=.9in]{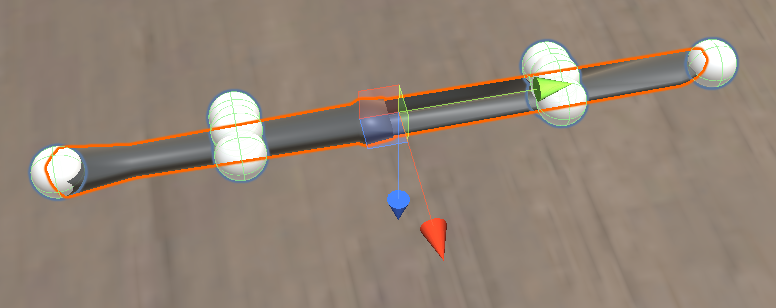}
\caption{\label{fig:grasp-points}Grasp points on a cup and knife.}
\end{center}
\vspace{-6mm}
\end{wrapfigure}

For objects with rotational symmetry, we find all points $P$ on the surface equidistant from the extremes along the axis of symmetry, as well as the extreme points of the object along that axis.  For objects without rotational symmetry, we find those points $P$ on each component of the object that intersect the plane(s) perpendicular to the plane of reflectional symmetry (see Fig.~\ref{fig:grasp-points}). The closest one of these points to the position of the agent's hand ($w$) is taken to be the targeted point of interaction with the object.  

From this point we calculate the maximum amount each finger ($f$) can bend toward the object without intersecting its bounds, take this distance of the fingers to wrist joint and add it to/subtract it from the object's extremities in both directions along all 3D major axes ($extents_{x,y,z}$).

\vspace{-20mm}
\begin{center}
\begin{equation*}
$$$\forall f \in F$ $\argmin\limits_{m\angle fwp}$ $\mid\vec{fw}\mid$ $\times$ cos($m\angle fwp$) $\pm$ $extents_{x,y,z}$$$
\end{equation*}
\end{center}
\vspace{-8mm}

The angle of the fingers at this point becomes a {\it grasp pose}, or hand position needed to activate the {\it grasping} affordance (that is, to exploit the {\it graspability}) of that object at that position.  Each possible grasp pose comes with an associated affordance semantics of the form $\lambda y.grasp(obj,with(y))$ where $y$ represents the associated iconic gesture, which is as yet unfilled.  Diana then demonstrates each inferred possible grasp pose to the human and asks for a preference:

\begin{dialogue}
	\speak{Human} The plate.
	\speak{Diana} Okay.
	\direct{
		\textsc{Diana} points to the plate.
	}

	\medskip
	\direct{
		\textsc{Human} makes ``claw down'' gesture.
	}
	\speak{Diana} Should I grasp it like this? \direct{
		\textsc{Diana} grasps plate from beneath.
	}
	
	\medskip
	\direct{
		\textsc{Human} makes ``thumbs down'' gesture.
	}
	\speak{Diana} How about like this? \direct{
		\textsc{Diana} grasps plate from side.
	}
	\speak{Human} Yes.
	\speak{Diana} {\bf Is there a gesture for that?}
	
	\medskip
	\direct{
		\textbf{\textsc{Human} makes ``grasp plate'' gesture.}
	}
\end{dialogue}

\begin{wrapfigure}{l}{.5\textwidth}
\vspace{-6mm}
\begin{center}
\includegraphics[width=.5\textwidth]{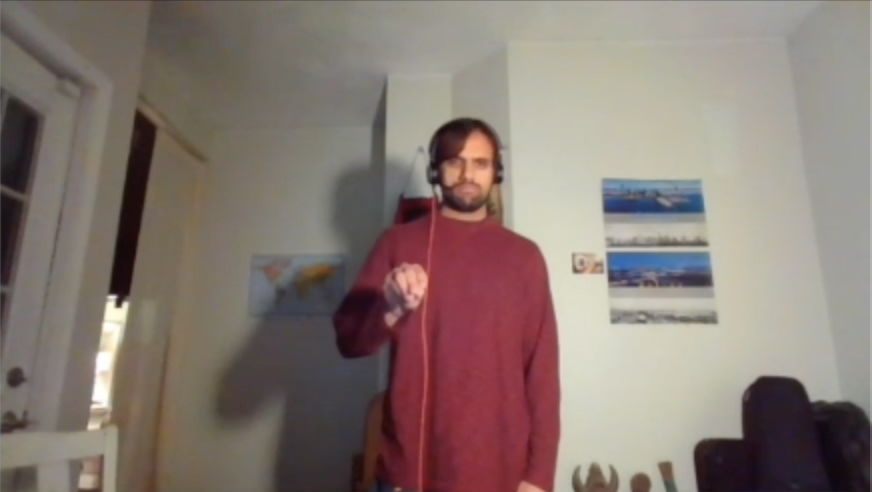}
\caption{\label{fig:plate-grab}Iconic {\it plate} gesture for ``grasp the plate.''}
\end{center}
\vspace{-6mm}
\end{wrapfigure}

Now, with the affordance semantics available to be filled, the visual features of the novel gesture the human makes are fed into a random forest classifier trained over 2048-dimensional feature vectors derived from the annotated video data used to train the default gesture recognizer.  The novel gesture is situated in the feature space of the 34 known gestures (plus any novel gestures previously learned).  That new vector value is applied to the outstanding variable in the affordance semantics generated through the interaction to this point.  The result represents an operationalization of $grasp(x)$ where $x$ is the object requiring novel exploitation of its affordances to grasp it.  This operationalized predicate is then propagated down to any other events that use [[{\sc grasp}]] as a subevent over the object $x$.  This now allows the human to instruct the agent to grasp an object using the correct pose, with a single visual cue, as in Fig.~\ref{fig:plate-grab}. Furthermore, the avatar can subsequently be instructed to perform any actions that subsume grasping a plate.

\begin{figure}[h!]
\centering
\includegraphics[width=.75\textwidth]{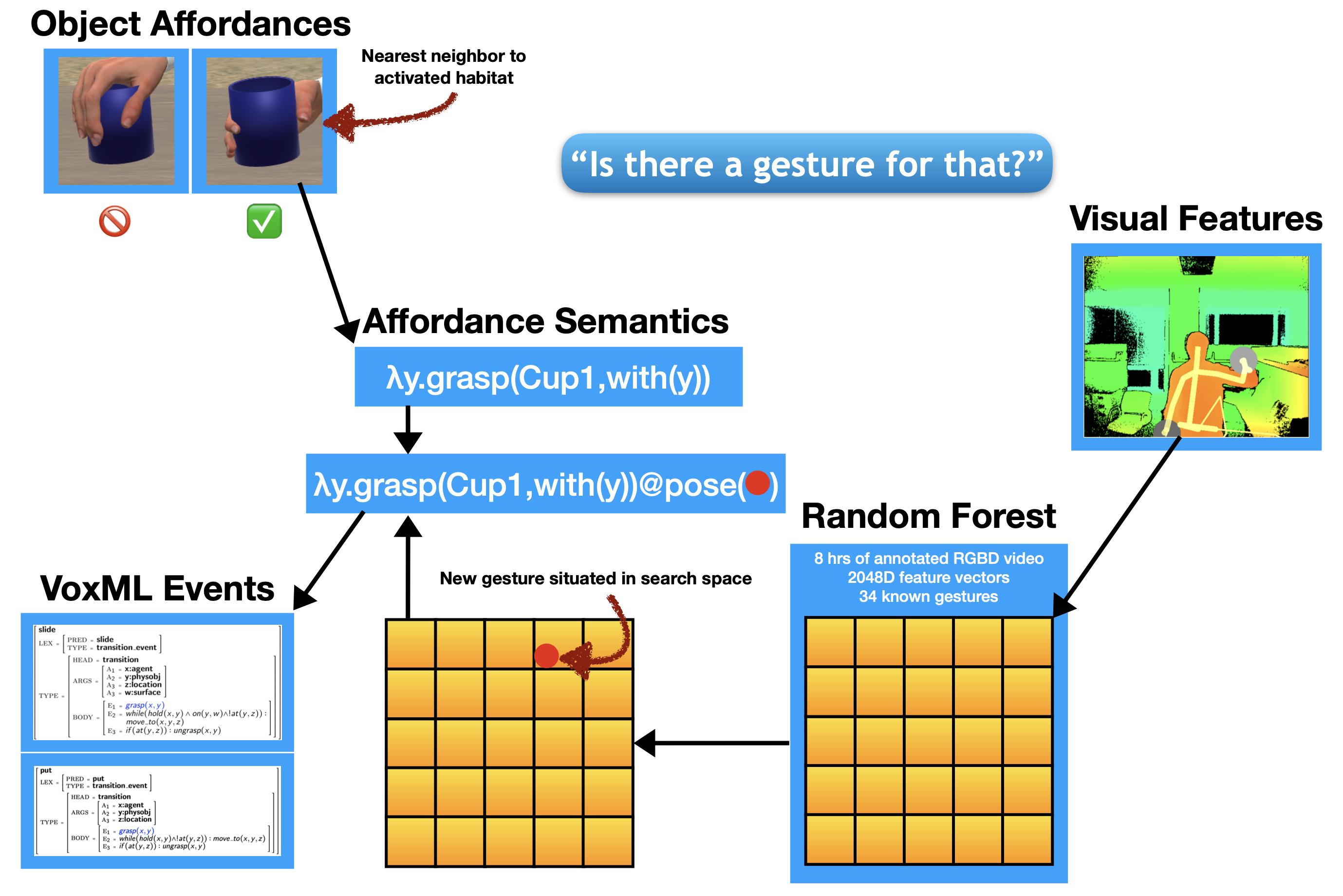}
\caption{Neurosymbolic pipeline for learning interactions with object affordances (Sec.~\ref{sec:affordances}).
}
\label{fig:afffordance-learning}
\end{figure}

Fig.~\ref{fig:afffordance-learning} gives the neurosymbolic learning pipeline for object affordances and accompanying actions.  Because the learned object affordance is propagated down to other events that contain the associated action, we can fill in other action sequences with this association using a continuation-passing style semantics \citep{krishnaswamy2019multimodal}.  For example, extending the dynamic event structure developed in \cite{PustMosz:2011}, the VoxML encoding of the event [[{\sc slide\_to}]] can be represented as in \ref{slide}. This is a derived event composed from the activity  [[{\sc slide}]] and the directional PP [[{\sc to\_loc}]] \citep{pustejovsky2014generating}. 

\begin{enumerate}[(1)]
\item  $grasp(e_1,{\mbox{\sc ag}},y);{\mbox {\bf \/ while}}(hold({\mbox{\sc ag}},y)\wedge on(y,{\mbox{\sc surf}})\wedge \neg at(y,{\mbox{\sc loc}})),move\_to(e_2,{\mbox{\sc ag}},y,{\mbox{\sc loc}}));$\\
${\mbox{\bf if}}(at(y,{\mbox{\sc loc}}),ungrasp(e_3,{\mbox{\sc ag}},y))$
\label{slide}
\end{enumerate}


Therefore, if the agent encounters a [[{\sc slide}]] action with an outstanding variable ($\lambda y.slide(y,loc)$), and the human supplies a gesture denoting $grasp(plate)$, then the agent can directly lift $grasp(plate)$ to the slide action and apply the argument $plate$ to $y$: $\lambda y.slide(y,loc)@plate \Rightarrow$ $slide(plate,loc)$.  {\bf while}$(C,A)$ states that an activity, $A$, is performed only if a constraint, $C$, is satisfied at the same moment. Here, should something cause the agent to drop the object or should the agent lift the object off the surface, the constraint, and therefore the overall [[{\sc slide}]] action will cease and remain incomplete.

\section{Learning Structure and Novel Configurations}
\label{sec:configs}

Configurations can be instantiated by exploiting an object's affordances to create relations between it and other objects.  For instance, exploiting the containment affordance of a cup (i.e., the structural properties of a cup that allow it to contain other items) by putting a spoon in it results in a ``spoon-in-cup'' configuration that can be realized in multiple ways under constraints (e.g., handle-up or handle-down, but not typically lying flat).  In this sense, complex configurations are the result of {\it composed affordances}.

This is a far more complex problem than learning affordances over single objects.  Naively, affordance composition of $k$ objects with $m$ affordances each runs in $O(m^k)$ time due to the need to check every enumerated affordance of an object against every enumerated affordance of every other object in an Allen-style composition table \citep{allen1983maintaining}, so for training and testing a model we back off to a simple Blocks World domain in order to reuse all affordances across all objects.  Even so, we encounter two problems:
\begin{enumerate}
\item Different configurations imply different sets of constraints.  For example, when laying a conventional place setting at a table, the plate goes in the center; putting it anywhere else creates something that is not a  ``place setting.''  When filling a moving truck with boxes or building a staircase out of blocks, there is simply a space to be filled or a configuration to be created, in lieu of very specific object placements.
\item Representative data is required for training.  The one-shot gesture learning featured in Sec.~\ref{sec:affordances} uses random forests over 2048-dimensional feature vectors trained over annotated RGBD video data, but no such equivalent data exists for this affordance composition/configuration problem, forcing reliance on smaller data sources and concomitant algorithms.
\end{enumerate}

\cite{krishnaswamy2018lrec} presents results of a user study wherein 20 naive users interacted with the Diana system.  They were instructed to build a three-step staircase out of six equal-sized, uniquely-colored blocks, and told that the system could understand gestures and spoken English, but were not given a specific vocabulary.  When the user was satisfied with the results, the task was considered complete.  Three users failed to complete the task in the allotted time, so the study produced 17 sample staircases of diverse configurations wherein the placements of individual blocks varied, stacks of blocks were not perfectly aligned, and staircases pointed in either applicable direction (see Fig.~\ref{fig:staircases}).  Therefore, any learning algorithm must be able to infer commonalities across this small and noisy sample.  This learning procedure is detailed in \cite{krishnaswamy2019combining} and can be analogized to common parts of a neural NLP pipeline using symbolic representations extracted from the situated simulated environment.

\begin{wrapfigure}{r}{.5\textwidth}
\vspace{-8mm}
\begin{center}
\includegraphics[height=.8in]{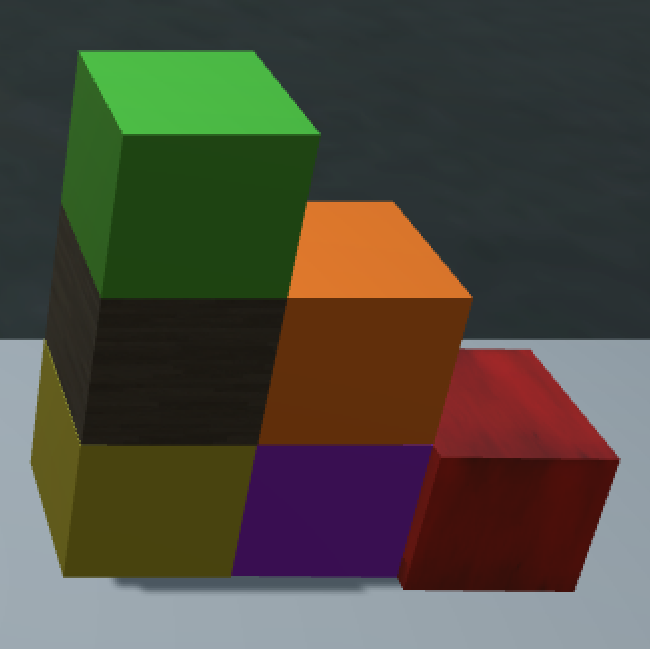}
\includegraphics[height=.8in]{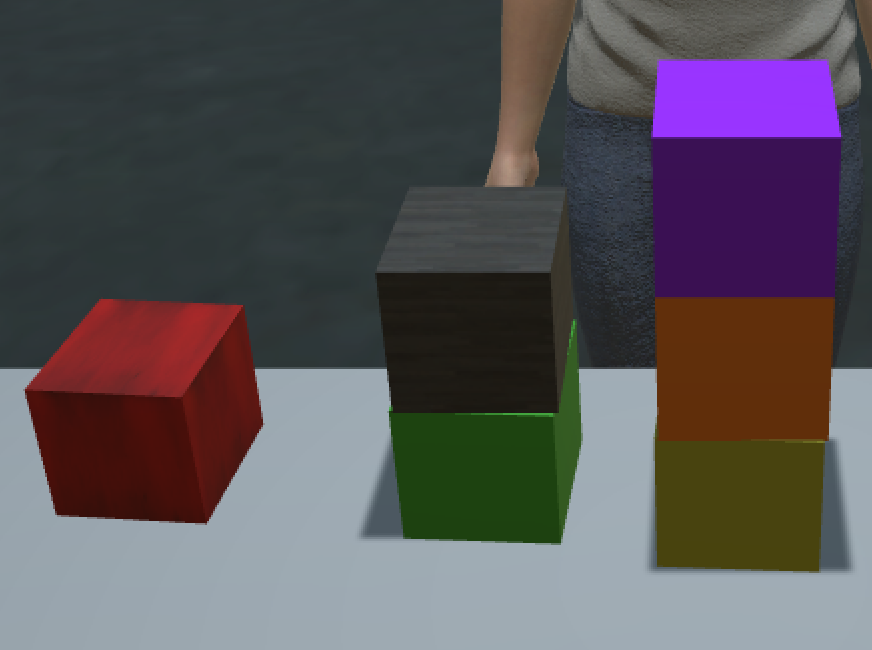}
\includegraphics[height=.8in]{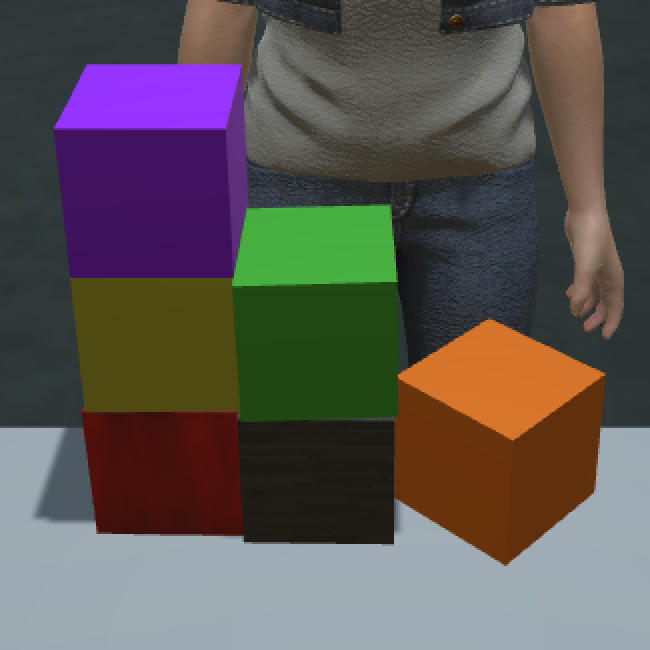}
\caption{\cite{krishnaswamy2018lrec}: sample user-constructed staircases.}
\label{fig:staircases}
\end{center}
\vspace{-6mm}
\end{wrapfigure}

The data we gathered is sparse enough that machine learning over the raw coordinates of blocks will fail to converge.  Therefore we made use of situated grounding to extract {\it qualitative} relations between objects.  We used subsets of the Region Connection Calculus RCC8 \citep{randell1992}, and the Ternary Point Configuration Calculus \citep{moratz2002qualitative} with implementations taken from QSRLib \citep{gatsoulis2016qsrlib}, which allow grounded, denser representations of the relation sets that make up a structure or configuration.

\begin{figure}[h!]
\centering
\includegraphics[width=.75\textwidth]{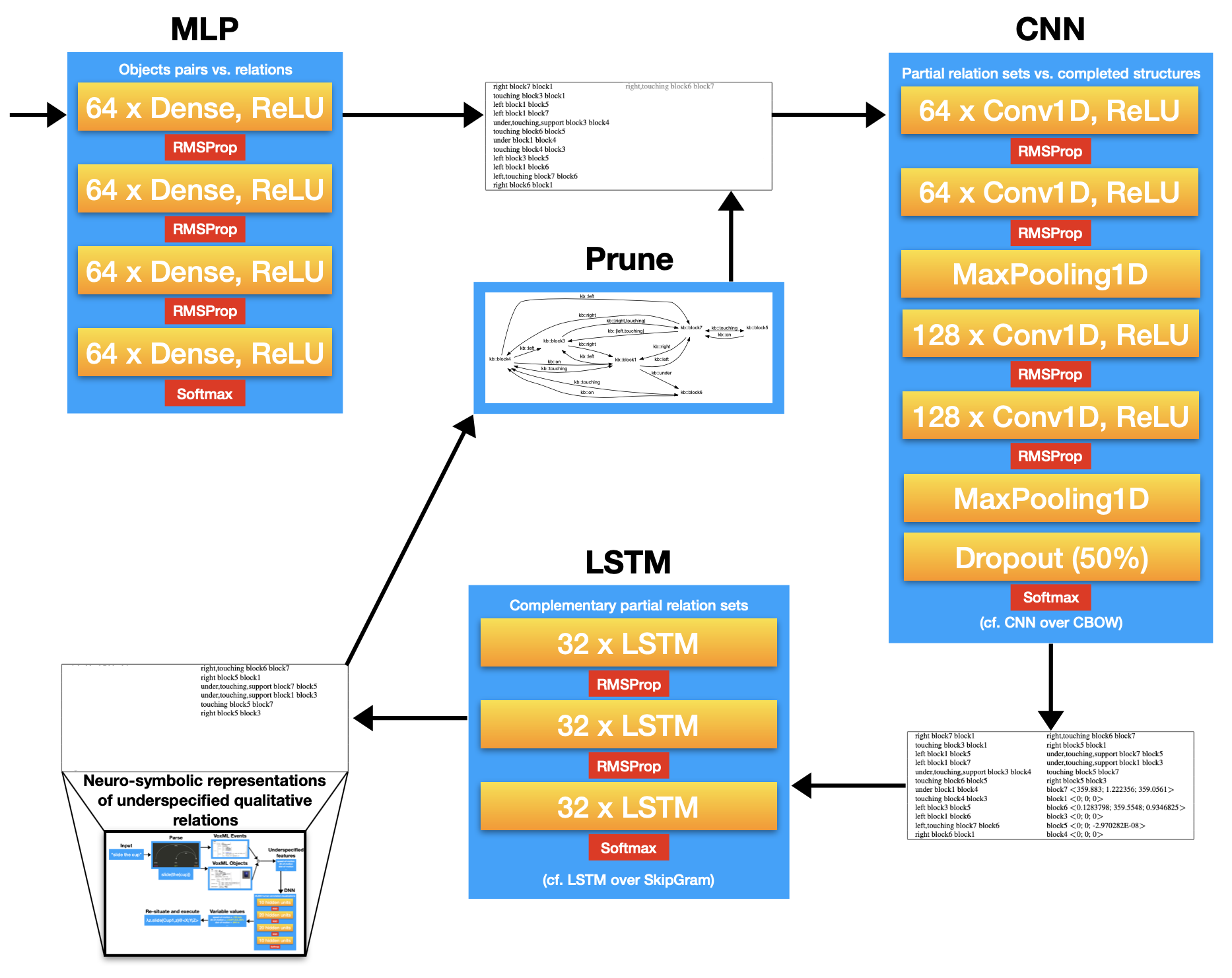}
\caption{Neurosymbolic pipeline for learning structural configurations.}
\label{fig:structure-learning}
\end{figure}

Fig.~\ref{fig:structure-learning} shows the neurosymbolic learning pipeline for structures and configurations given situated grounding.  To sample from the training data and to avoid artificially introducing some of the inferences the learning pipeline is intended to pick up itself, the first move is generated using a multi-layer perceptron that takes in a random pair of objects and outputs a relation to create between them.  This generates a partial relation set that is fed into the relation-classifying CNN.  The CNN should output a known, completed structure that contains that partial relation set.  This can be analogized to a continuous bag-of-words model over a vocabulary of relations.  This output and the partial relation set representing the current state are fed into the LSTM network, which outputs a set of move options that would complement the current partial relation set and be most likely to complete the structure predicted by the CNN.  This is similar to a sequence construction task or an LSTM trained over a Skip-Gram model.  Each move option presented by the LSTM is a symbolic representation of a qualitative spatial relation to be created between two blocks.  As such, the specifics of where to place the blocks in Cartesian coordinate space must be filled in somehow, and so the operationalization can then use the pipeline for generating multimodal simulations (Fig.~\ref{fig:mmsim}) to complete this process.  The move options are then pruned to select a single best move that is then enacted, adding relations to the current state's partial set.  This process repeats until the objects are used up.

\clearpage
\subsection{Validation}
\label{ssec:config-validation}

\begin{wrapfigure}{r}{.55\textwidth}
\vspace{-6mm}
\begin{center}
\includegraphics[width=.25\textwidth]{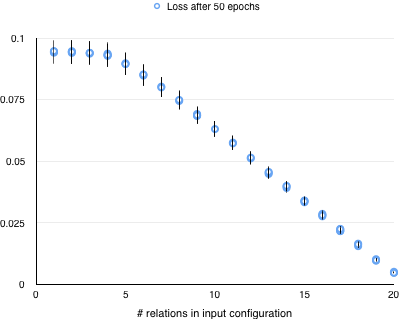}
\includegraphics[width=.25\textwidth]{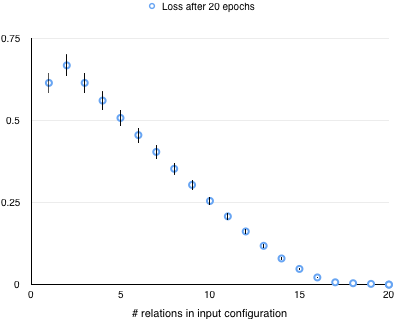}
\caption{\label{fig:loss}CNN (L)/LSTM (R) loss vs. input length.}
\end{center}
\vspace{-6mm}
\end{wrapfigure}

The CNN and LSTM portions of the learning framework above rely on a notable assumption: that when only a few relations are instantiated, a known example and complementary relation set will be almost random but as more relations are instantiated, the prediction of both neural networks becomes more accurate.

To validate this assumption, we sampled increasingly large relation sets from the training data, fed them through both networks, and measured the cross-entropy loss after 50 epochs (for the CNN) and 20 epochs (for the LSTM).  As expected (Fig.~\ref{fig:loss}), the prediction of both nets becomes more accurate as relation input size increases, and approaches almost 0 once 20 relations are instantiated, representing, in most cases, a complete structure from the training data.

\subsection{Results}
\label{ssec:config-results}

\begin{wrapfigure}{r}{.5\textwidth}
\vspace{-6mm}
\begin{center}
\includegraphics[height=.7in]{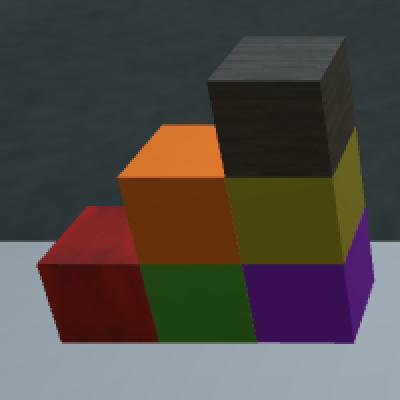}
\includegraphics[height=.7in]{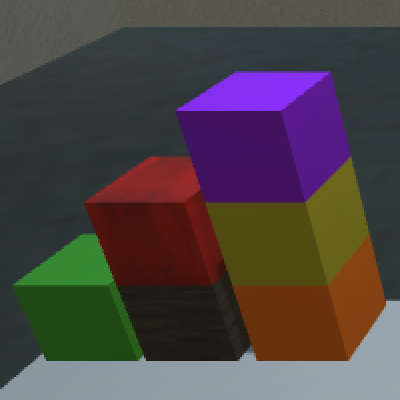}
\includegraphics[height=.7in]{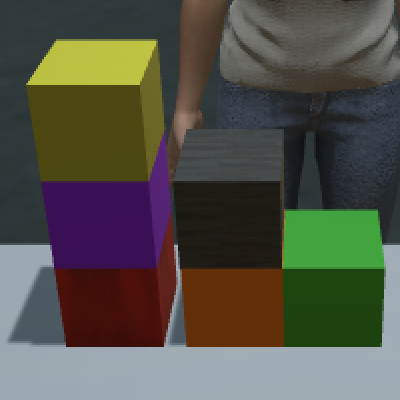}
\includegraphics[height=.7in]{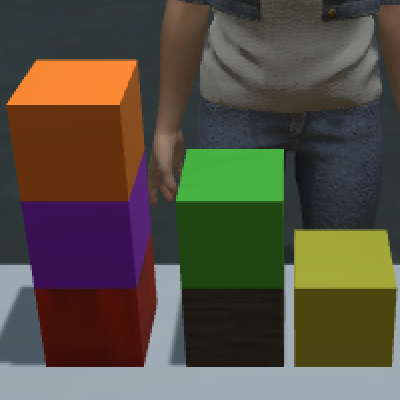}
\caption{Sample generated staircases.}
\label{fig:sample-structs}
\end{center}
\vspace{-6mm}
\end{wrapfigure}

Fig.~\ref{fig:sample-structs} shows sample generated staircases displaying desired inferences.  To prune the output of the two stacked neural nets to choose a single move to execute, we assessed 5 heuristic loss functions: random chance as a baseline, Jaccard distance to measure the shared presence or absence of a spatial relation \citep{jaccard1912distribution}, Levenshtein distance to measure the {\it count} of shared spatial relations \citep{levenshtein1966binary}, a graph-matching algorithm called {\sc spire} \citep{mclure2015extending}, and a Levenshtein distance-pruned version of {\sc spire}.  We generated 50 novel instances of staircases in total; 5 from each trained model, and had a set of evaluators rate them all from 0-10, based on the question ``How much does the structure shown resemble a staircase?''  Images were viewed in random order to minimize sequential biases across evaluators.  {\sc spire} was the most successful heuristic loss function; the average evaluator score for staircases generated using {\sc spire} was {\bf 5.8313}, compared to {\bf 2.0375} for random chance, and {\bf 4.7188} for the next-highest-performing heuristic, the Levenshtein distance-pruned graph matcher.

Situated grounding within a multimodal simulation provides methods for inferring the composition of structures and configurations from small sample sizes, specifically the ability to translate from specific coordinates to qualitative relations and back within a neurosymbolic model.  This addresses the {\it generation} aspect of novel concept learning for an AI agent.

\subsection{Semantic Grounding}
\label{ssec:config-grounding}

Supplementary to generation is the ability to recognize and classify instances of novel concepts.  Within VoxWorld this is addressed as a constraint satisfaction problem.  Since we are primarily concerned with learning the habitats within which a novel structural configuration exists, and the afforded behaviors that those habitats enable, we provide the system with the components that make up a novel structure, and allow it to infer what affordances of those components are being used in the novel structure, and which satisfy the constraints in the learned and generated samples.

\begin{wrapfigure}{r}{.15\textwidth}
\begin{center}
\vspace{-6mm}
\includegraphics[width=.15\textwidth]{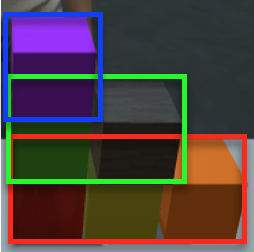}
\caption{Staircase components.}
\label{fig:components}
\vspace{-6mm}
\end{center}
\end{wrapfigure}

In \cite{krishnaswamy2019situated} this is approached from many angles, including weighted constraint satisfaction or as a partially-observable Markov decision process (cf. \cite{lee2018monte}).  One promising avenue, using both the advantages of qualitative representation within situated grounding and mapping cleanly to the learning pipeline described above, is a qualitative constraint network (QCN).  We use a variant of the CONACQ.2 algorithm outlined in \cite{mouhoub2018learning} where the language consists of spatial relations from the training data and generated examples (replacing Allen Temporal Relations \citep{allen1983maintaining}), the background knowledge is transitive closure rules over the relation vocabulary, the learning strategy is the pipeline outlined above, and the result is encoded in VoxML.

\begin{algorithm}[H]
\SetKwData{Left}{left}\SetKwData{This}{this}\SetKwData{Up}{up}
\SetKwFunction{Union}{Union}\SetKwFunction{FindCompress}{FindCompress}
\SetKwInOut{Input}{input}\SetKwInOut{Output}{output}
\SetAlgoLined
\Input{Relation language $\mathcal{B}$, background knowledge $K$, strategy $\mathcal{L}$}
\Output{Acquired constraint set $\mathcal{S}$}
	$\mathcal{S}$$\leftarrow$\{\}; $converged$ $\leftarrow$ $false$\;
	\While{$\neg$$converged$}{
		$q$ $\leftarrow$ $QueryGeneration(\mathcal{B}, \mathcal{S}, K, \mathcal{L})$\;
		\eIf{$q$ = $nil$}{
			$converged$ $\leftarrow$ $true$\;
			}{
			\eIf{$Answer(q)$ $>$ $threshold$}{
				$\mathcal{S}$ $\leftarrow$ $\mathcal{S}$ $\wedge$ $\bigwedge_{c \in K(q)}$ $c\vdash$ $\mathcal{S}$\;
			}{
				$\mathcal{S}$ $\leftarrow$ $\mathcal{S}$ $\wedge$ ( $\bigvee_{c \in K(q)}$ $c\nvdash$ $\mathcal{S}$)\;
			}
		}
	}
	return $\mathcal{S}$\;
	\caption{\label{alg:conacq}Adapted CONACQ.2 for habitat and affordance inference.}
\end{algorithm}

\begin{figure}[h!]
\centering
{\tiny\def\baselinestretch{1.1}
$\avmplus{\att{staircase}\\
	\attval{type}{\avmplus{
		\attvaltyp{head}{assembly[1]}\\
		\attvaltyp{components}{{\color{red}base[2]},{\color{green}step[3]*},{\color{blue}top[4]}}
	}}\\
	\attval{habitat}{\avmplus{
        		\attval{Intr}{[5]
			\avmplus{
				\attvaltyp{base}{$align({\color{red}[2]},\mathcal{E}_X)$}\\
				\attvaltyp{up}{$align(vec(loc({\color{blue}[4]})-loc({\color{red}[2]})),\mathcal{E}_Y)$}
			}
		}
	}}\\
	\attval{afford\_str}{\avmplus{
		\attvaltyp{A$_1$}{$H_{[5]} \rightarrow [put(x,on([1]))]part\_of(x,[1])$}\\
		\attvaltyp{A$_2$}{$H_{[5]} \rightarrow [put(x,on({\color{red}[2]}))]part\_of(x,{\color{green}[3]})$}\\
		\attvaltyp{A$_3$}{$H_{[5]} \rightarrow [put(x,left \vee right \vee $}\\
		\attvaltypnoatt{\;}{$touching({\color{red}[2]}) \wedge \neg on({\color{red}[2]})]extend(x,{\color{red}[2]})$}\\
		\attvaltyp{A$_4$}{$H_{[5]} \rightarrow [put(x,left \vee right \vee $}\\
		\attvaltypnoatt{\;}{$touching({\color{green}[3]}) \wedge \neg on({\color{green}[3]})]extend(x,{\color{green}[3]})$}
	}}
}$
\def\baselinestretch{1.9}}
\caption{Acquired constraints describing a staircase.}
\label{fig:conacq}
\end{figure}

In the query generation phase of Algorithm $\ref{alg:conacq}$, we back off to the LSTM used in step 3 of the learning strategy $\mathcal{L}$ (Fig.~\ref{fig:structure-learning}), and make use of the attention vectors to generate the queries.  Thereafter, if the answer to the query is $true$ above a given threshold in the example under examination, it is added to the constraint set, otherwise it is removed if present.

Fig.~\ref{fig:conacq} shows a sample result, and that we can successfully extract certain constraints that describe not just the staircase shown, but an abstract staircase.  These constraints include: the steps ascending to either the left {\it or} right (A$_3$, A$_4$), that placing an object on the base or step creates a new (non-base or -step) tier (A$_3$, A$_4$), or that putting something on the base makes it part of the step (A$_2$).  This provides at least some of the components of a semantic model for the new object.  When asked about a ``staircase,'' the agent now has semantics for a decontextualized reference that can be reproduced and adjusted without having to retrain the purely numerical model.  Thus situated grounding allows probing of models in a tractable way by examining qualitative relations and acquired constraints.

\section{Transfer Learning of Object Properties and Linguistic Description}
\label{sec:transfer}

Through correlating cross-modal representations (e.g., the visual features of a gesture and an embodied action in 3D space, or the composition of qualitative relations and a structural label or its semantics), or correlating cached symbols with neural representations (e.g., a relation or action label and a set of specifically quantitative operationalizations), situated grounding serves as a platform for improving sample efficiency through reuse.  Therefore, it should also facilitate transferring knowledge gained from solving one problem and applying it to another situation.  Situatedness is particularly useful for transfer learning, because similar concepts often exist in similar situations (cf. analogical generalization, a la \cite{forbus2017extending}).

Associating affordances with abstract properties or symbolic labels informs the way that the entities that possess those affordances can be discussed.  In Sec.~\ref{sec:intro}, we discussed the inability of unimodal language understanding systems to answer the simple question ``what am I pointing at?''  While situated grounding provides a solution to linking linguistic terms to entities sharing the agent's co-situated space, the agent can still only discuss these entities if she knows the appropriate terms for them.  If an agent encounters a new object that she doesn't know the name of, she can discuss it in terms of ``this one'' or ``that one,'' but cannot decontextualize the reference with a lexical label.  Transfer learning provides a way to give the agent a way of talking about a novel item by comparing it to known items.

We focus here on transfer learning of affordances.  Since similar objects typically have similar habitats and affordances (e.g., cylindrical items with concavities often serve as containers), it is worth investigating whether such properties can be transferred from known objects to novel objects that are observed to have similar associated properties.
The method we use is termed {\it affordance embedding}.  This follows an intuition similar to the Skip-Gram model in natural language processing \citep{mikolov2013efficient}, or the masked language model of BERT \citep{devlin2018bert}, but exploits the linkage between affordances and objects present in a situated grounding model like VoxWorld.  We are experimenting with two architectures: a 7-layer MLP and a 4-layer CNN with 1D convolutions, both trained over habitat-affordance pairs taken from our vocabulary of known objects.  For instance, a habitat-affordance pair for a [[{\sc cup}]] voxeme might be ($H_{[2]}$ = [{\sc up} = $align(Y,\mathcal{E}_Y)$, {\sc top} = $top(+Y)$], $H_{[2]}$ $\rightarrow$ $[put(x,in(this))]contain(this,x)$) (gloss: {\it the cup's Y-axis is aligned upward with the Y-axis of the embedding space, and if something is put inside the cup, the cup contains that thing}).  The network outputs collocation probabilities for every  individual habitat-affordance pair in the vocabulary.  Subsequently, for each possible action the agent may take with an object (e.g., grasp, pick up, move, slide, put on, etc.), the system queries the learned affordance embeddings, {\it excluding} those affordances that include the particular action in question.  This restates the answer to a query, e.g., ``describe the appropriate habitat for {\it grasping} an object'' in terms of {\it other} actions that can be taken in that habitat, and the habitat is matched to other objects that share that habitat. This is effectively a second-order collocation.

For example, if the agent comes across an unfamiliar object that appears to share the $H_{[2]}$ = [{\sc up} = $align(Y,\mathcal{E}_Y)$, {\sc top} = $top(+Y)$] (upward alignment) habitat of [[{\sc cup}]], she can infer that it might be grasped in a similar way.  Fig.~\ref{fig:transfer} shows this process enacted through dialogue.  In frame 1, the human points to a new object (recognizable as a bottle, but Diana has no label associated with it).  In frame 2, Diana says ``I don't know``---reflecting the semantic gap in her vocabulary---``but I can grasp it like a cup''---reflecting the information about it that she is able to infer from its habitats and affordances, which gives her a way to talk about this object with her human partner.  In frame 3, the human says ``grab it,'' and Diana demonstrates her inferred method of grasping, using grasp pose calculation as described in Sec.~\ref{sec:affordances}.

\begin{figure}[h!]
\centering
\includegraphics[height=1.3in]{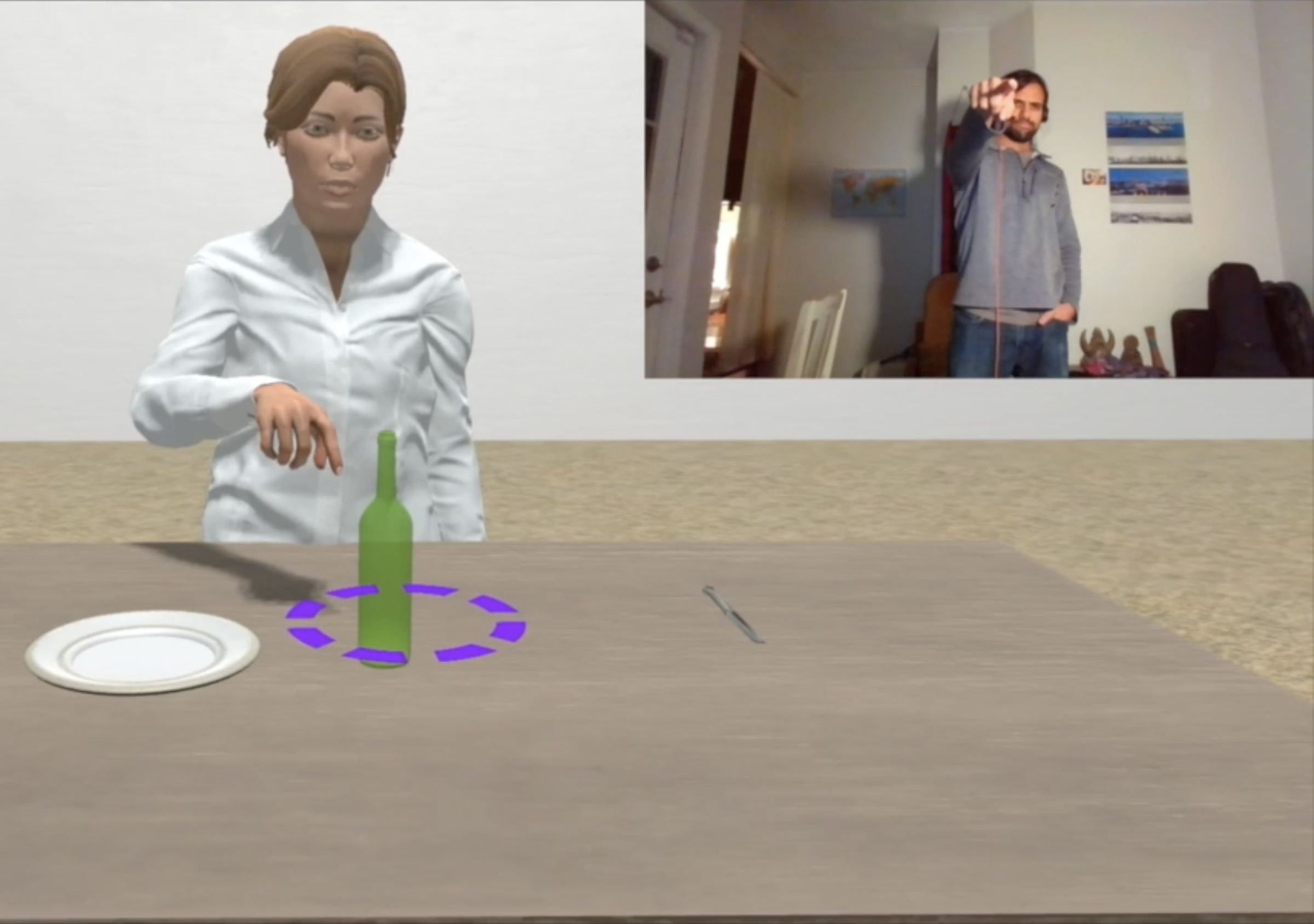}
\includegraphics[height=1.3in]{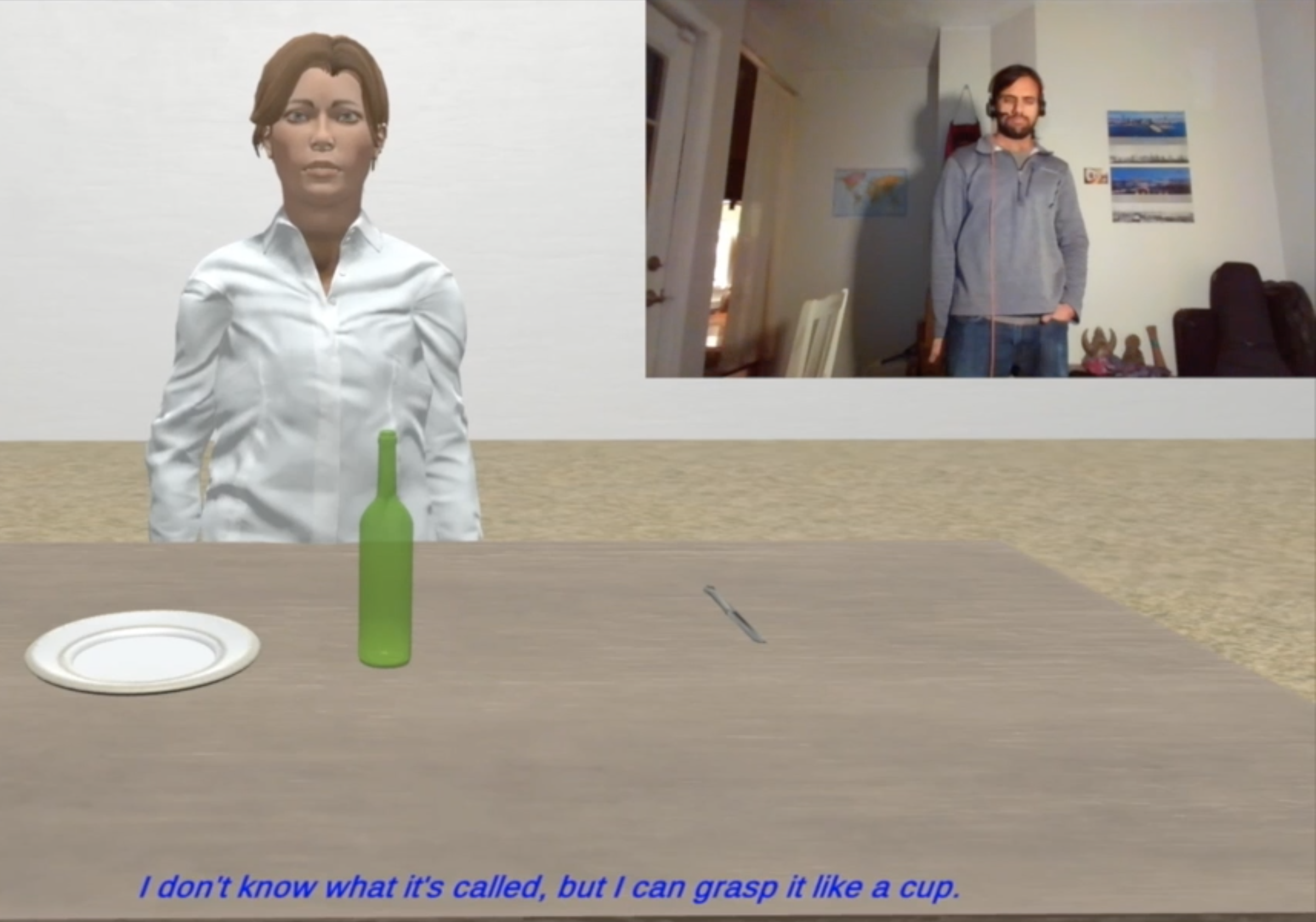}
\includegraphics[height=1.3in]{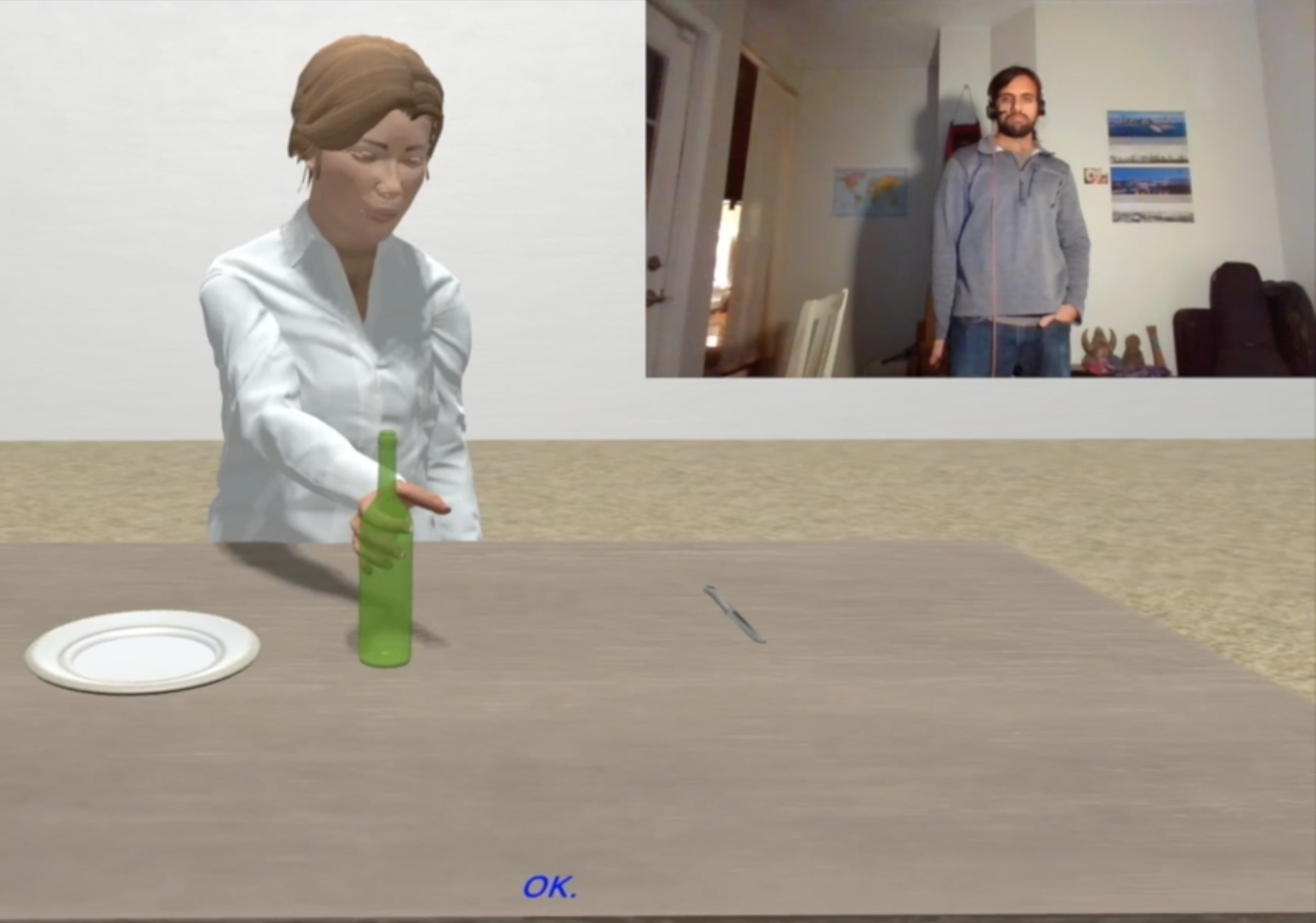}
\caption{\label{fig:transfer}1) Human: ``What is that?''; 2) Diana: ``I don't know, but I can grasp it like a cup.''; 3) Human: ``Grab it.'' + resultant grasp.}
\end{figure}

\section{Conclusions}
In this paper, we hope to have demonstrated that the notion of situatedness goes well beyond  visually grounding a text or a concept to an image or video; rather, it involves embedding the linguistic expression and its grounding within a multimodal semantics.  This approach provides for environmentally-aware models that can be validated; each additional modality provides an orthogonal angle through which to validate models of other modalities.  It provides many methods of encoding context both quantitatively and qualitatively, and provides a model to accommodate both neural and symbolic representations.  The diverse types of data available through a situated grounding platform are adaptable to different tasks with novel types of network architectures, with less data overhead than typical machine learning.  As such, in an era where mainstream AI tends toward increasingly large  datasets and bigger models involving more and more parameters, with concomitant costs in energy and resource usage, such a platform provides a sustainable way toward more powerful AI.
 
\begin{acknowledgements} 
\noindent
We would like to thank our collaborators at Colorado State University and the University of Florida for their longtime collaboration on developing the Diana interactive agent.  We would also like to thank the reviewers for their helpful comments.  This work was supported by the US Defense Advanced Research Projects Agency (DARPA) and the Army Research Office (ARO) under contract \#W911NF-15-C-0238 at Brandeis University.  The points of view expressed herein are solely those of the authors and do not represent the views of the Department of Defense or the United States Government.  Any errors or omissions are, of course, the responsibility of the authors.
\end{acknowledgements} 

\vspace{-0.25in}

{\parindent -10pt\leftskip 10pt\noindent
\bibliographystyle{cogsysapa}
\bibliography{References}

}


\end{document}